\documentclass[review]{elsarticle}

\usepackage[left=3cm, right=3cm]{geometry} 
\usepackage{parskip}
\makeatletter
\def\ps@pprintTitle{%
 \let\@oddhead\@empty
 \let\@evenhead\@empty
 \def\@oddfoot{\leftline{Manuscript under review}}%
 \let\@evenfoot\@oddfoot}
\makeatother

\usepackage{breakurl}

\usepackage[utf8]{inputenc} 
\usepackage[T1]{fontenc}    
\usepackage{hyperref}       
\usepackage{url}            
\usepackage{booktabs}       
\usepackage{amsfonts}       
\usepackage{nicefrac}       
\usepackage{microtype}      
\usepackage{graphicx}       
\usepackage{amsmath}
\usepackage{mathtools}
\usepackage{algorithm}
\usepackage{algorithmic}
\usepackage{subcaption}

\usepackage{comment}
\usepackage{xcolor}
\usepackage{wrapfig}

\begin{document}
\begin{frontmatter}

\title{Predictive coding feedback results in perceived illusory contours in a recurrent neural network}

\author[mymainaddress]{Zhaoyang Pang}
\author[mymainaddress]{Callum Biggs O'May}
\author[mymainaddress]{Bhavin Choksi}

\author[mymainaddress,mytertiaryaddress]{Rufin VanRullen\corref{mycorrespondingauthor}}
\cortext[mycorrespondingauthor]{Corresponding author}
\ead{rufin.vanrullen@cnrs.fr}

\address[mymainaddress]{CerCO, CNRS UMR5549, Toulouse}
\address[mytertiaryaddress]{ANITI, Toulouse}

\begin{abstract}
Modern feedforward convolutional neural networks (CNNs) can now solve some computer vision tasks at super-human levels. However, these networks only roughly mimic human visual perception. One difference from human vision is that they do not appear to perceive illusory contours (e.g. Kanizsa squares) in the same way humans do. Physiological evidence from visual cortex suggests that the perception of illusory contours could involve feedback connections. Would recurrent feedback neural networks perceive illusory contours like humans? In this work we equip a deep feedforward convolutional network with brain-inspired recurrent dynamics. The network was first pretrained with an unsupervised reconstruction objective on a natural image dataset, to expose it to natural object contour statistics. Then, a classification decision layer was added and the model was finetuned on a form discrimination task: squares vs. randomly oriented inducer shapes (no illusory contour). Finally, the model was tested with the unfamiliar ``illusory contour'' configuration: inducer shapes oriented to form an illusory square. Compared with feedforward baselines, the iterative ``predictive coding'' feedback resulted in more illusory contours being classified as physical squares. The perception of the illusory contour was measurable in the luminance profile of the image reconstructions produced by the model, demonstrating that the model really ``sees'' the illusion. Ablation studies revealed that natural image pretraining and feedback error correction are both critical to the perception of the illusion. Finally we validated our conclusions in a deeper network (VGG): adding the same predictive coding feedback dynamics again leads to the perception of illusory contours.

\end{abstract}

\end{frontmatter}

\section{Introduction}
\begin{figure}[h]
    \centering
    \includegraphics{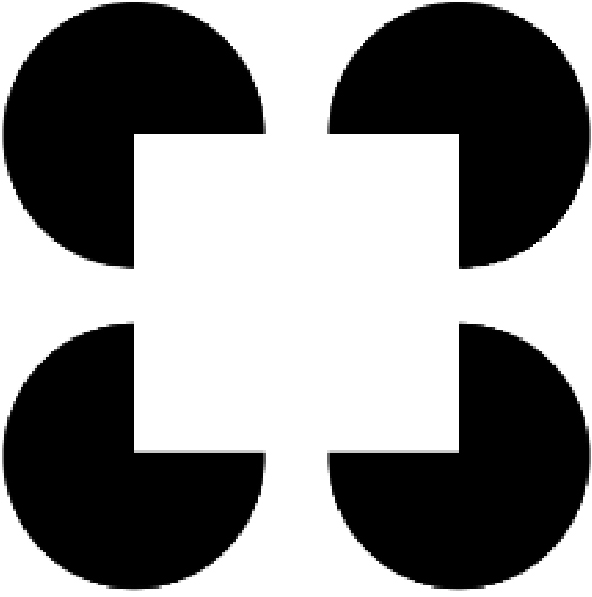}
    \caption{\textbf{Kanizsa Square illusion} }
    \label{fig1-kanizsa}
\end{figure}

The human visual system is remarkably versatile. It is capable of accurately recognizing objects from different angles, over different distances, under distortion, and even under occlusion. Despite this, it can easily be fooled by simple visual illusions. Figure~\ref{fig1-kanizsa} shows an example of Kanizsa illusory contour (or subjective contour)~\cite{kanizsa1955margini,kanizsa1976subjective}. When observing such an image, most humans perceive the presence of a square, despite the only shapes present being the pacman-shaped inducers. In fact, in addition to perceiving edges of the illusory square, humans tend to perceive the interior of the induced shape as being brighter than the exterior, despite their being the same. These two phenomena illustrate two salient features of illusory figures -- sharp illusory edges in regions of homogeneous luminance, and a brightness enhancement in the figure~\cite{Schumann1901-SCHBZA-3,spillmann1995phenomena,parks2001rock}. There are many examples of illusions like this, demonstrating that the human visual system does not always accurately perceive stimuli. Such systematic misperceptions reflect underlying neural constraints and provide insight into the complex structure of visual cortex~\cite{changizi2008perceiving}. Visual illusions have thus been used as a probe for understanding visual processing~\cite{eagleman2001visual,gori2016visual}. 
Alongside traditional neuroscientific approaches to studying visual perception, advances in machine learning have led to new avenues in understanding the mechanisms of visual processing. By developing artificial models of human vision we can investigate neuroscientific principles and reduce the need for experimentation. In particular, computational models allow for rapid, iterative experimentation and development. In fact, the origin of modern artificial neural networks (the backbone of much modern artificial intelligence or AI) is in computational models of neurons~\cite{mcculloch1943logical}. Throughout the history of AI development, researchers have taken inspiration from the brain. A clear example of this is the development of convolutional neural networks (CNNs), which were a critical step in the computer vision revolution~\cite{fukushima1982neocognitron, lecun1989backpropagation}. Inspired by the hierarchical structure of the brain, CNNs limit the spatial extent of the neuronal receptive fields, resulting in retinotopic feature maps much like the early visual system. In the last decade, the explosion of deep learning research has led to massive improvements in object recognition, even reaching super-human performance~\cite{He_2015_ICCV}. However, the rapidly changing research landscape has produced many technical developments which are not always directly compatible with neuroscience. As a result, the relationship between human vision and computer vision has become less clear. A striking example of this is learning through error backpropagation, which uses a biologically implausible global error signal~\cite{lillicrap2020backpropagation}. Recent attempts have been made to identify more biologically plausible learning rules~\cite{whittington2017approximation, millidge2020predictive,ahmad2020gait,lee2015difference}. Many studies have also sought to investigate the similarities and differences between modern computer vision and human vision~\cite{khaligh2014deep, yamins2014performance, cichy2016comparison, wen2018neural}. Such work highlights the dual benefits of computational neuroscience, which can both provide guidance for designing computer vision algorithms, and contribute to our understanding of brain functioning.

Despite all this, a recent report~\citep{Baker2018DeepCN} shows that convolutional neural networks do not perceive illusory contours in the same way humans do. In their study, a feedforward CNN -AlexNet~\citep{krizhevsky2012imagenet} - was trained to perform a thin/fat contour discrimination task on real and illusory contours. In both cases, the network could correctly classify the images, but the authors demonstrated that the representations of the illusory contours in the CNN do not resemble those of human observers. Specifically, they employed a so-called classification image technique~\citep{gold2000deriving} to give insight into which regions in a given image are important for the illusory formation. For human subjects, the region between inducers is critical to the perceptual decision~\citep{gold2000deriving}. The feedforward neural network, on the other hand, failed to interpolate illusory contours between inducers -- instead, it relied on the orientation of inducing elements to make its decisions~\citep{Baker2018DeepCN}. That is, instead of adopting a global processing strategy as humans did, the CNN mainly appeared to rely on local features. In a related study, Kim et al.~\citep{kim2020neural} directly examined the representations in intermediate layers of a feedforward neural network. The authors computed the cosine similarity between the network's representations of physical and illusory contours, and showed that they are more similar to each other than to control non-illusory shapes. However, this representation similarity measure is only an indirect indication of the perception of illusory contours, since it does not explain \textit{which} features are similar between the illusory contours and the physical contours. It could be, for example, that the similarity is related to the presence of correctly-oriented corners, rather than to the contours between them. Thus, it may still be true that feedforward CNNs do not perceive illusory contours, but instead largely base their decisions on local features. Other evidence shows that feedforward CNNs can perform poorly on tasks which depend explicitly on global processing, like long-range spatial dependencies \cite{linsley2018learning} or object recognition under occlusion~\citep{Spoerer2017recurrence}. In summary, it would seem that the visual perception of feedforward artificial neural networks is more dependent on local processing than human perception.

Perceptual discrepancies between artificial and biological networks highlight fundamental differences in their underlying structures as well as performed computations. Feedforward CNNs only roughly mimic the visual system, and particularly the feedforward pass through the visual pathway, while feedback connections, although abundant in the brain, are often ignored. Notably, physiological evidence shows that illusory contour perception may rely on feedback connections from V2 to V1~\cite{lee2001dynamics,pak2020top} or in higher-level regions~\cite{cox2013receptive,pan2012equivalent}. In this respect, the generation of illusory contours could be interpreted within the predictive coding framework~\cite{notredame2014visual,raman2016predictive}, a popular computational theory of feedback processing introduced in neuroscience by Rao and Ballard~\cite{rao1999predictive}. This theory posits that, in a hierarchical system, each layer tries to predict the activity of the layer below, and the prediction errors are used to update the activations. In fact, predictive coding could serve as a unified computational principle for a variety of sensory systems including auditory~\cite{kumar2011predictive}, olfactory~\cite{zelano2011olfactory} and visual sensation~\cite{mumford1992computational,nour2015perception}. In the field of machine learning, a number of recent works have also applied the predictive coding framework to modern deep learning networks. Boutin et al. demonstrated that a predictive coding network with a sparsity constraint learns similar receptive fields to neurons in V1 and V2, and provided evidence that the feedback helps the network perform contour integration~\cite{boutin2021sparse}. Lotter et al. showed that the predictive coding framework can be used to learn representations of pose and motion in video streams~\cite{lotter2017deep}. Importantly, their model was only trained with unsupervised objectives, arguably closer to human learning than standard supervised learning methods. In addition to the biological plausibility of predictive coding networks, other works have demonstrated that these networks can show improved robustness to random or adversarial noise~\cite{choksi2020, huang2020neural, chalasani2013deep}. Wen et al. studied whether predictive coding can result in improved classification of clean images~\cite{wen2018deep}, but since their network was not trained to minimize reconstruction error, it did not fulfill the predictive coding objective as described by Rao and Ballard (reduction of reconstruction errors over timesteps). A further discussion of the limits of the work of Wen et al.~\cite{wen2018deep} can be found in~\cite{choksi2020}. 

Given both the significant role of feedback connections in biological illusory contour perception, and the potential for predictive coding as neuroscientific inspiration for feedback architectures in deep learning models, we hypothesized that a feedback neural network implementing predictive coding recurrent dynamics may perceive illusory contours in the same way humans do. The recent work of Lotter et al.~\citep{lotter2018neural} strengthens this hypothesis. They investigated the responses of PredNet~\citep{lotter2017deep}, a predictive coding network, to illusory contours. They compared the response properties of model units to neuronal recordings in the primate visual cortex and showed comparable response dynamics in the presence of illusory contours~\citep{lotter2018neural}. \textcolor{black}{Unlike the current work, which inspects the network's behavioural and perceptual responses, the Lotter et al. work approached the topic entirely at the level of neuronal activations.} Although their research supports the idea of a similarity in contour representations between the PredNet artificial layers and the biological cortex, \textcolor{black}{they do not read off network decisions, or inspect reconstructions of the network. Thus, they do not argue that the network is human-like on a behavioural or perceptual level.  As with the above-mentioned Kim et al. paper~\cite{kim2020neural}, it could be that the similarities in the representations are due to local low-level features (e.g. the inward-facing corners) rather than the illusory contours. Thus, it} remains unclear whether their model truly ``perceives'' illusory shapes like humans do.

In the current study, we designed a deep predictive coding neural network according to the algorithm previously devised by our group~\cite{choksi2020}. We used a relatively small network, consisting of three stacked autoencoders, intended to roughly \textcolor{black}{mimic the hierarchical structure of the} early visual cortex in the primate brain, since neural correlates of illusory contours have been found to involve visual areas V1, V2 and V4~\cite{von1984illusory,grosof1993macaque,pak2020top,pan2012equivalent,cox2013receptive}. The network had both generative (image reconstruction) and discriminative (image classification) capabilities. Therefore, we could not only directly check whether the network indeed ``sees'' illusory contours, by examining its reconstruction images from generative feedback connections, but also measure the network's ``behavioral'' performance, i.e. the discriminative readout values indicating whether or not it detects an illusory shape. In addition, in order to equip our model with sufficient contour knowledge, we propose and verify that it is necessary to pretrain the neural network on natural images; this is in line with evidence that human contour integration and grouping is strongly tied to the statistics of the natural world \cite{geisler2009contour}. Finally, we extend this approach to a deeper feedforward network (VGG) and demonstrate that this network sees the illusory contours too. In summary, we report that our predictive coding feedback networks tend to process illusory contours in a similar way to humans.

\section{Materials and Methods}

\subsection{Architecture}

We construct a three-layer hierarchical stacked autoencoder with 3 feedforward encoding layers $e_{n}$ ($n\in\ {1,2,3} $) and 3 corresponding feedback decoding layers $d_{n-1}$ (see Figure \ref{fig-architecture} and Table \ref{table:1}).  \textcolor{black}{This is a conventional network architecture for learning a lower-dimensional latent representation of a high-dimensional input space~\cite{kingma2013auto}}. When considering only the encoding layers, the network can be viewed as a standard feedforward convolutional neural network. To guide the implementation of the feedback connections, we follow the principles of ``predictive coding'' as introduced by Rao and Ballard~\citep{rao1999predictive}: in the hierarchical network, the higher layers try to predict the activity of the lower layers and the errors made in this prediction are then used to update their activity. For a given input image, we initiate the activations of all encoding layers with a feedforward pass. Then over successive recurrent iterations (referred to as timesteps $t$) we update the decoding and encoding layer representations using the following equations: 
\begin{figure}[t]
    \centering
    \includegraphics{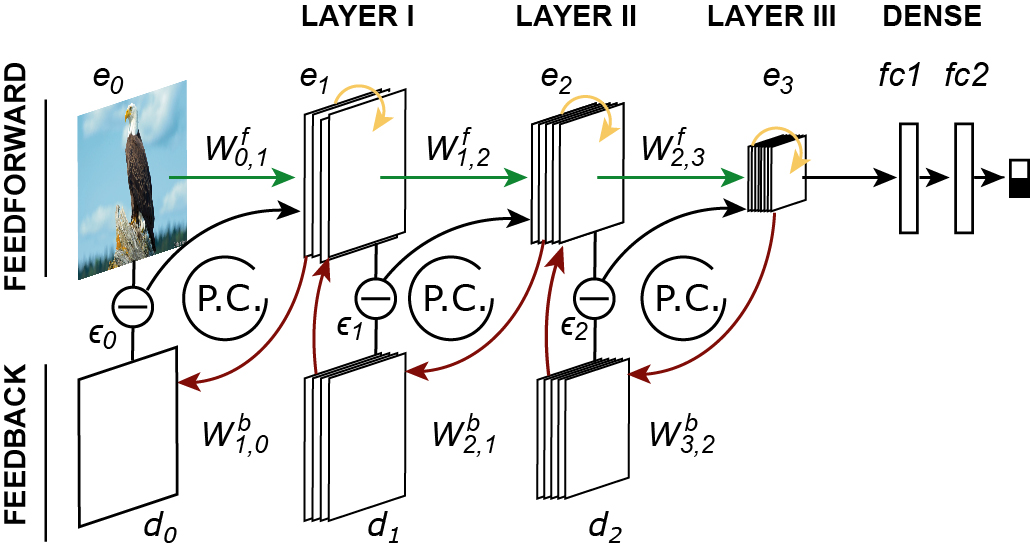}
    \caption{\textbf{Network architecture.} The architecture consists of a main body and a classification head (or dense layers). For the main body, the predictive coding strategy is implemented in stacked autoencoders, with three feedforward encoding layers ($e_{n}$) and three generative feedback decoding layers ($d_{n}$). Reconstruction errors ($\epsilon_{n}$) are computed and used for the proposed predictive coding updates which are denoted by ``P.C.'' loops. Dense layers are added on top of the structure to implement a binary classification task.}
    \label{fig-architecture}
\end{figure}

\begin{table}[h!]
\centering
\begin{tabular}{c c}
    \hline
    layers & parameters \\
    \hline
    $e_{1}$ & $\big[ Conv(3,5,2) \big]_{+}$ \\
    $e_{2}$ & $\big[ Conv(128,5,2) \big]_{+}$ \\
    $e_{3}$ & $\big[ Conv(128,5,2) \big]_{+}$ \\
    $d_{0}$ & $\big[ Conv(3,5,2) \big]_{Sig.}$ \\
    $d_{1}$ & $\big[ Conv(128,5,2) \big]_{+}$ \\
    $d_{2}$ & $\big [ Conv(128,5,2) \big]_{+}$ \\
    $fc1$ & $\big[ W(2048,256) \big]_{+}$ \\
    $fc2$ & $\big[ W(256,128) \big]_{+}$ \\
    \hline
\end{tabular}
\caption{\textbf{Table of parameters.} Each encoding layer is a combination of a convolution layer and a ReLU nonlinearity with parameters Conv(channels, kernel size, stride). \textcolor{black}{All convolutions have padding 4}. Decoding layers consist of a deconvolutional layer with ReLU non-linearity, except for layer $d_{0}$, which uses a Sigmoid activation function, in order to compare with the input picture with pixel values ranging from 0 to 1. After flattening the output of the last convolutional layer, and going through a batch normalization function, two dense layers with a structure of Weight(in features, out features) project to the binary decision layer.}
\label{table:1}
\end{table}

\begin{equation}
\begin{split}
\label{eq:pc_equation}
    \textcolor{black}{d_{n}(t)} &= \textcolor{black}{\big[ W^{b}_{n+1,n}e_{n+1}(t) \big]_{+}} \\
    \textcolor{black}{e_{n}(t+1)} &= \textcolor{black}{\beta \big[ W^{f}_{n-1,n}e_{n-1}(t+1) \big]_{+} + \lambda d_{n}(t) + (1 - \beta - \lambda) e_{n}(t) - \alpha\nabla\epsilon_{n-1}(t)} 
\end{split}
\end{equation}
where $W^{f}_{n-1,n}$ denotes the feedforward weights connecting layer $n-1$ to layer $n$, and $W^{b}_{n+1,n}$ denotes the feedback weights from layer $n+1$ to $n$. \textcolor{black}{$\epsilon_{n-1}(t)$ is the reconstruction error for layer $n-1$: the mean squared error between the representation $e_{n-1}$ and the corresponding prediction $d_{n-1}$:}

\begin{equation}
\textcolor{black}{\epsilon_{n-1}(t)=\|e_{n-1}(t)-d_{n-1}(t)\|_{2}^{2}}
\end{equation}

\textcolor{black}{Then $\nabla\epsilon_{n-1}(t)$ denotes the gradient of the error at layer $n-1$ with respect to the activations in layer $e_n$. That is, $\nabla\epsilon_{n-1}(t)$ is the vector of partial derivatives of the error with respect to each element of $e_n$ (and thus has the shape of $e_n$), so that the $i$th element of $\nabla\epsilon_{n-1}(t)$ will be $\frac{\partial \epsilon_{n-1}(t)}{\partial e_{n}(t)_i}$. We use the PyTorch automatic differentiation package to calculate this one-step gradient.} The parameters $\beta$, $\lambda$ and $\alpha$ act as balancing coefficients for the feedforward drive, feedback error correction, and feedforward error correction terms respectively, and they are treated as hyperparameters of the network (for the present experiments, except where otherwise noted, these hyperparameter values were fixed to $\beta=0.2$, $\lambda=0.1$ and $\alpha=0.1$ as this was found to be sufficient for producing the illusion). \textcolor{black}{We can also rewrite Equation 1 by grouping the terms for each hyperparameter, to more clearly illustrate the connection with the Rao and Ballard formulation of predictive coding:}

\begin{equation}
\begin{split}
\label{eq:pc_equation3}
    \textcolor{black}{e_{n}(t+1)} &= \textcolor{black}{\beta \big[ W^{f}_{n-1,n}e_{n-1}(t+1)\big]_+ + (1 - \beta) e_{n}(t) + \lambda \big[d_{n}(t)  -  e_{n}(t)\big]_{+}  - \alpha\nabla\epsilon_{n-1}(t)} 
\end{split}
\end{equation}

\textcolor{black}{This clarifies that the $\lambda$ hyperparameter is the weight for the term $d_{n}(t)  -  e_{n}(t)$, which is exactly the gradient of $\epsilon_n(t)$ with respect to $e_n(t)$. Thus we see that the $\lambda$ and $\alpha$ terms are exactly the two error terms in Rao and Ballard's formulation. A fuller demonstration of the relationship between the two formulations can be found in the Appendix.}

Since the error $\epsilon_{n-1}$ is an average over the whole representation in the layer below (see Eq. \ref{eq:pc_equation2}), as the number of units in the representation increases, the error term variance will tend to shrink to~0. Additionally, the retinotopic nature of the convolution operation (with connectivity restricted to a local neighborhood) means that most connections between the layers are 0, and thus have 0 gradient. These combined effects result in small gradients relative to the layer's activations, so to counteract this we re-scale the gradient terms according to the layer and kernel sizes. \textcolor{black}{Specifically, for each layer, we calculate for the layer below $K$=channels × width × height and $C$=channels × kernel size, and then multiplicatively scale the gradient term by a factor of $K/\sqrt C$. This directly balances out the two effects discussed. See Appendix for full details.}

\textcolor{black}{To reflect the systematic comparison between decoding and encoding layers, we set $e_0$ as our input images. Thus, our updating rule in Eq.~\ref{eq:pc_equation} is only applied to $e_n$($n\in\ {1,2,3} $) and $e_0$ will remain constant over timesteps. In addition, for the last layer $e_3$ in our model, there is no feedback, so we ignore the corresponding term in the update equation.} All the weights $W$ are fixed during the updates defined by Eq.~\ref{eq:pc_equation}. They are optimized over successive batches of natural images and across all timesteps (see Training procedure) to minimize the total reconstruction error $L$ (Eq.~\ref{eq:pc_equation2})--an unsupervised objective in accordance with the principles of the predictive coding theory. In Eq.~\ref{eq:pc_equation2}, $N$ is the number of layers (here, $N = 3$). \textcolor{black}{We note that Equation \ref{eq:pc_equation} leads to updates which approximately reduce the loss in Equation \ref{eq:pc_equation2} over timesteps, as in both Rao and Ballard's formulation and Whittington and Bogacz \cite{whittington2017approximation}.}

\begin{equation}
\label{eq:pc_equation2}
    L = \sum_{t} \sum_{\mathclap{n=0}}^{N-1} \epsilon_{n}(t) = \sum_{t} \sum_{\mathclap{n=0}}^{N-1} \|e_n(t)-d_n(t)\|_{2}^{2}
\end{equation}

Intuitively, each of the four terms in Eq.~\ref{eq:pc_equation} contributes different signals to a layer: (i) the feedforward term (controlled by parameter $\beta$) provides information about the (constant) input and changing representations in the lower layers, (ii) the feedback error correction term (parameter $\lambda$), hereafter referred to simply as ``feedback'', guides activations towards their representations from the higher levels, thereby reducing the reconstruction errors over time, (iii) the memory term helps to retain the current representation over successive timesteps, and (iv) the feedforward error correction term (controlled by parameter $\alpha$) corrects representations in each layer such that their next prediction better matches the preceding layer, also contributing to the reduction in reconstruction errors. Together, the feedback and feedforward error correction terms fulfill the objective of predictive coding as laid out by Rao and Ballard \cite{rao1999predictive}.

\subsection{Training procedure}
The network's training includes two stages: pretraining and finetuning (both using 10 timesteps for inference). The pretraining (over 150 epochs) was conducted in an unsupervised way with a reconstruction objective (see Eq.~\ref{eq:pc_equation2}), wherein both feedforward and feedback convolution weights were optimized over the CIFAR100 natural images dataset~\citep{krizhevsky2009learning}. This was done to learn a hierarchy of relevant features to describe each natural image, as well as the corresponding generative pathway to reconstruct images from their features. 

For the second stage, we added a 3-layer classification head to the network (consisting of three fully-connected layers), and finetuned all parameters of the network on the custom dataset presented below with a supervised binary cross-entropy loss. The weights of the whole network were finetuned for 25 epochs, after which it was tested with a new validation set. We performed three distinct pretrainings (each with different randomly initialized weights), and then each of these pretrainings was used to finetune 3 distinct networks. The reported test results are averaged over the resulting 9 networks.

\textcolor{black}{During both pretraining and finetuning we use the backpropagation-through-time (BPTT) approach, which unrolls the network through time, and then backpropagates errors. We use the PyTorch automatic differentiation for efficiency and simplicity. We use the ADAM optimizer \cite{kingma2014adam}, a stochastic gradient method with automatically-adapting learning rates. This algorithm automatically uses momentum (and we use the default PyTorch hyperparameters of 0.9, 0.999) to estimate the first and second order moments of gradients. We use an initial learning rate of $5e-5$.}

\textcolor{black}{When using BPTT, the computational requirements increase with the number of timesteps. Thus there is generally a balance required between the power of the model (where more timesteps allows for more complex computations) and feasibility. We chose 10 timesteps to produce meaningful dynamic trajectories while remaining within our computational constraints. We show in the Appendix that we still see reasonable illusory perception results with other timestep values.}

\subsection{Stimuli}
To test the perception of illusory contours we designed a custom dataset based on the Kanizsa illusion. We systematically generated stimuli consisting of an image with either a square, or four pacman-shaped inducers. The network was trained on a simple binary discrimination task: to classify each image as either a square, or pacman inducers. We generated four types of stimuli: (i) the real square; (ii) the illusory condition with all 4 inducers facing inwards (All-in); (iii) a control condition with all inducers facing outwards (All-out); and (iv) a random condition with random inducer orientation, but neither all in nor all out (Random). We finetuned the dataset for classification using only types (i) and (iv), and used (ii) and (iii) only at test time. In this way, the illusory shape is not part of the training set (out-of-sample), so the network cannot explicitly learn to categorise it either as a square or pacmen. The control condition was used to verify how the network reacts to stimuli which are out-of-sample but do not produce an illusion in humans.

\begin{figure}[t]
    \centering
    \includegraphics{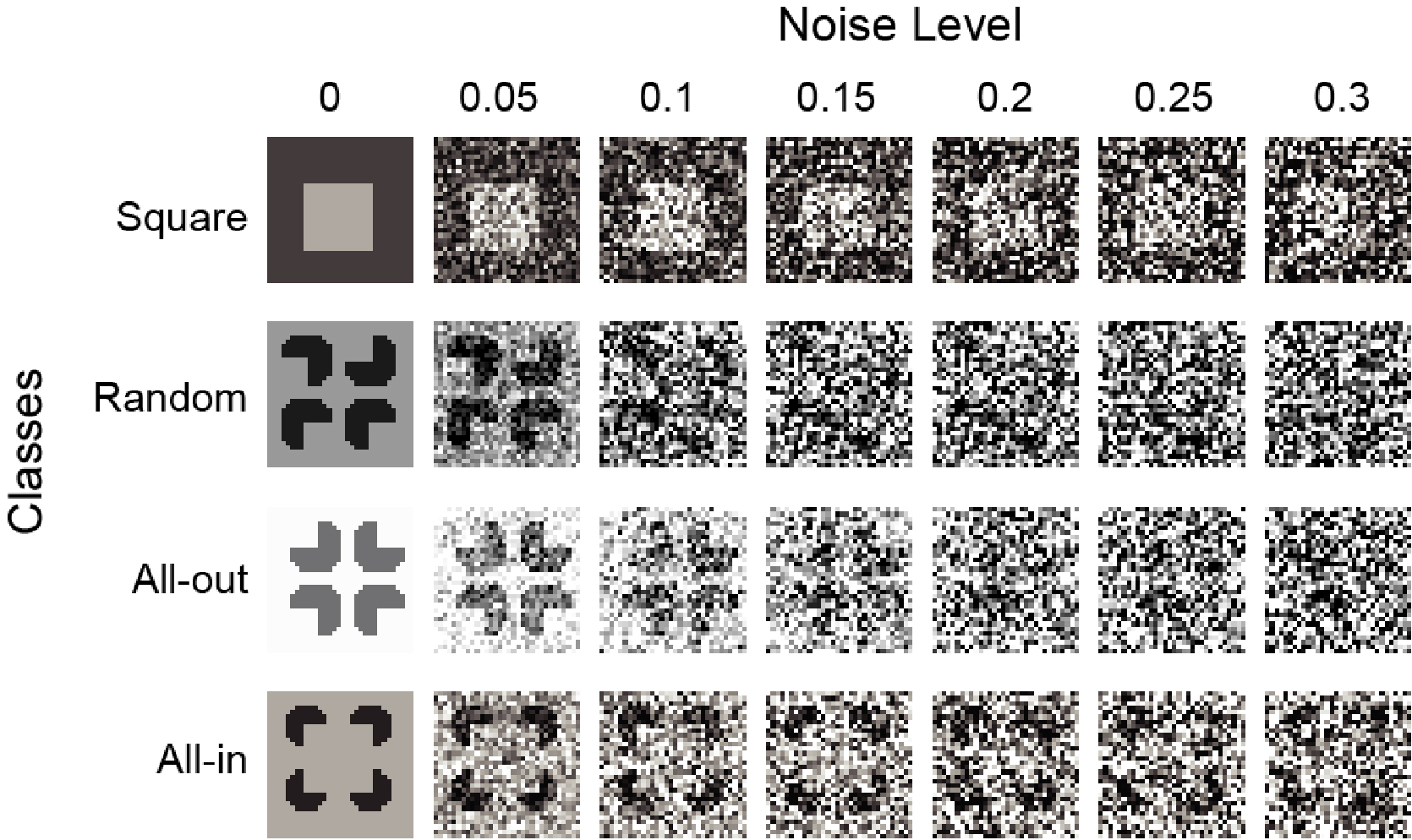}
    \caption{\textbf{Finetuning dataset.} Sample training and testing images for 7 different levels of Gaussian noise: Square, Random, All-out, and All-in.  All stimuli varied in luminance (both background luminance and inducer luminance), size (square sidelength or distance between inducers), and position. To enhance visualization, we displayed all seven levels of noises with each sample picture; while in practice, only one level of noise was randomly assigned to each stimulus configuration.}
    \label{fig3-Stimuli}
\end{figure}

Each image had a resolution of 32 × 32, consistent with the size of the CIFAR100 dataset used in the pretraining stage. The stimuli were designed to vary in luminance (for both the background and inducers, varying between 0 and 1), size, and position. This encourages the network to learn general rules and avoids overfitting through simple rote memorisation of the training set. For the same purpose, we also added Gaussian noise to each image (with variance randomly drawn from 7 pre-set levels varying from 0 to 0.3). Figure \ref{fig3-Stimuli} shows sample stimuli, with the different levels of noise. We generated a training set of 10,000 images, 5,000 each for the Square and Random classes, and another set of 2,500 images for validation. For testing, we generated 1200 images of each class for a total of 4,800 images.

\subsection{Feedforward baselines} 
The behaviour of the network at the first timestep (here $t=1$, after feedforward initialization of activations throughout the network, but before error correction updates are applied according to Eq~\ref{eq:pc_equation}) can be understood as a pure feedforward network. However, the network's features have been trained as part of the autoencoder blocks to optimize an unsupervised reconstruction objective -- not (or not only) a supervised classification objective as in standard feedforward CNNs. We thus also chose to compare our model to a pure feedforward network (FF) trained only with a supervised classification loss. For a direct comparison, we also trained this network on CIFAR 100, before finetuning it on the shapes. This network has exactly the same architecture and number of parameters as the predictive coding network when considered only at $t=1$, with the only difference being the learning objective.

However, this feedforward network has many fewer parameters overall than the full predictive coding network (around half). Thus, to directly compare to a feedforward network with roughly the same number of parameters, we also constructed two other feedforward networks: FF-C with more channels in each layer, and FF-K which uses a larger kernel size \cite{Spoerer2017recurrence}. Table \ref{table:2} compares the number of parameters for each network.

\begin{table}[h!]
\centering
\begin{tabular}{c c c c}
    \hline
    Models & Kernel Size & No. Channels & No. Parameters \\
    \hline
    FF & 5 × 5 & (3,128,128) & 1,403,620 \\
    FF-C & 5 × 5 & (3,172,172) & 2,248,684 \\
    FF-K & 7 × 7 & (3,128,128) & 2,199,268 \\
    PC & 5 × 5 & (3,128,128) & 2,232,679 \\
    \hline
\end{tabular}
\caption{\textbf{Comparison of parameters for feedforward networks.} PC is our predictive coding architecture. FF comprises only the feed-forward pass, but is trained with a supervised classification objective. FF-C is obtained by increasing the number of channels; FF-K by increasing the kernel size.}
\label{table:2}
\end{table}

\subsection{Code availability}
All the code, along with the pretrained models and stimulus datasets used in this work \textcolor{black}{is} available at:~\href{https://github.com/rufinv/Illusory-Contour-Predictive-Networks}{\nolinkurl{https://github.com/rufinv/Illusory-Contour-Predictive-Networks}}

\section{Results}

\subsection{Illusory contour perception}
\subsubsection{Shape classification}
After pretraining on natural images and fine-tuning on simple shapes, the network could discriminate between physical squares and randomly oriented inducers in any configuration (except the two critical configurations, All-in and All-out, which had not been seen during training). During testing, all four classes were presented to the network, with All-out (control class) and All-in (illusory contours) as novel stimulus configurations formed by familiar inducers. Our hypothesis was that when presented with the All-in configuration, the network would assign a higher probability to the square class than when presented with the All-out configuration.

\begin{figure}[t]
    \centering
    \includegraphics{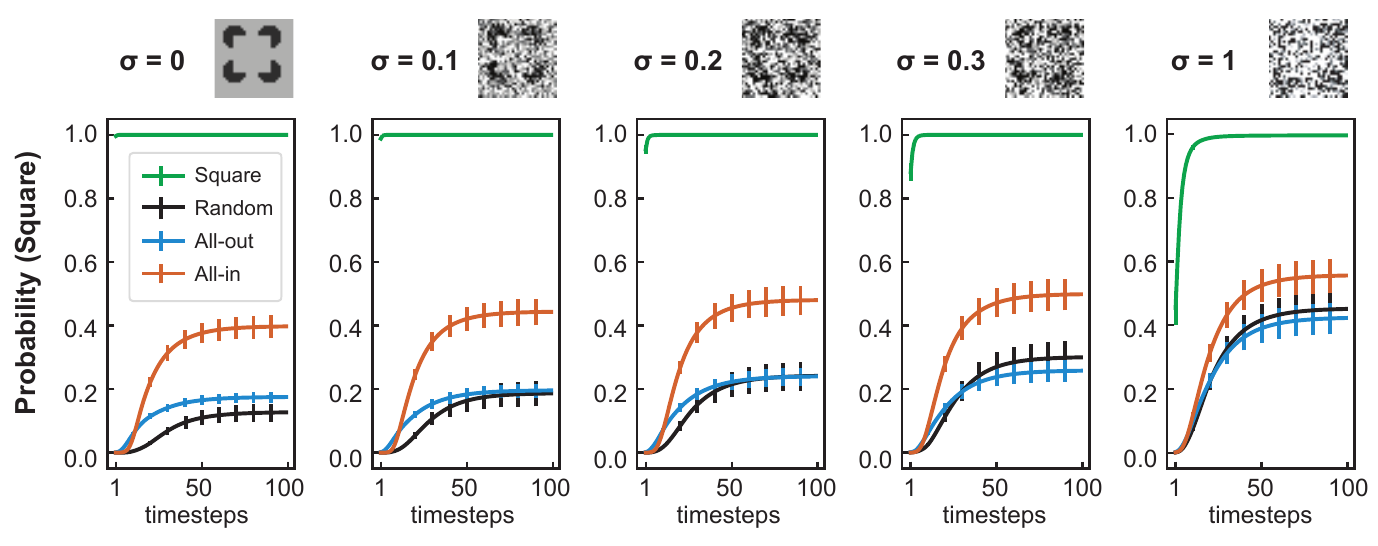}
    \caption{\textbf{Classification results.} The probability of ``square'' report over recurrent predictive coding iterations (timesteps). Each panel shows a different noise level -- for visibility, only three out of six levels of noise are shown here, in addition to the clean images ($\sigma = 0$). Feedback iterations increase the likelihood of ``square'' report, especially for the All-in (illusory contour) condition. Results are averaged over 9 networks and error bars represent standard error of the mean (SEM) across the networks.}
    \label{fig:decisions}
\end{figure}

For each class, we thus inspected the probability assigned by the network to the square category for each image (Figure \ref{fig:decisions}, results averaged over 9 networks). At timestep 1 (with only feedforward processing having taken place), the network appears to classify images based on low-level, local information, as all the pacmen-made patterns (Random, All-out, All-in) are recognized as non-square shapes. However, over timesteps, the network begins to recognize the All-in condition (the illusory contour) as a square, at a much higher rate than the All-out and Random conditions. After 50 timesteps, the average probability assigned by the network to the square class (an ``illusory contour perception'') increases by more than 40\% for the All-in condition, compared to around 20\% (depending on noise level) for the other conditions. Although this measure does not go above 50\%, we suggest that even humans would not actually categorize inducers as squares---despite `seeing' the illusion, we easily recognize that there is no actual square in the image.  As the noise level increases (Figure \ref{fig:decisions}, right), the likelihood of reporting a square for the All-in condition \textcolor{black}{stays roughly constant, whereas the likelihood for the All-out and Random conditions increases. This can be interpreted as the network becoming more `confused', as all probabilities are drifting together. The \textit{difference} between the illusory condition and the control conditions gets smaller as the noise level increases, but remains visible for high noise. This is consistent with previous work which demonstrated that humans perceive illusory contours even under noise \cite{gold2000deriving}.}
\begin{table}[h!]
\centering
\begin{tabular}{c c c c c}
    \hline
    Models & Square & No. Random & All-out & All-in \\
    \hline
    FF & 0.999 & 0.000 & 0.002 & 0.002 \\
    FF-C & 0.999 & 0.000 & 0.002 & 0.003 \\
    FF-K & 0.999 & 0.000 & 0.002 & 0.003 \\
    PC(t=1) & 0.981 & 0.000 & 0.001 & 0.000\\
    PC(t=100) & 1.000 & 0.000 & 0.201 & \textbf{0.478} \\
    \hline
\end{tabular}
\caption{\textbf{Comparison to feedforward networks.} The probability of square outcome for each class, for the feedforward baselines and the predictive coding network (including its initial timestep, which can also be viewed as a feedforward network, but with an additional unsupervised pretraining as part of a stacked autoencoder architecture). We see that the feedforward networks assign a square probability of almost 0 to all non-square classes, whereas the predictive coding network, by the last time timestep, has assigned a probability of nearly 0.5.}
\label{tab:ff_results}
\end{table}

We also compared our predictive coding network to the various feedforward baselines. Table \ref{tab:ff_results} shows the results. We clearly see that all feedforward networks assign a near-zero average square probability to both the All-in and All-out conditions, showing that they do not perceive the illusion at all. This further confirms the hypothesis that feedback connections are critical for the perception of illusory contours.

\subsubsection{Image reconstructions}
\begin{figure}[htbp]
    \centering
    \includegraphics{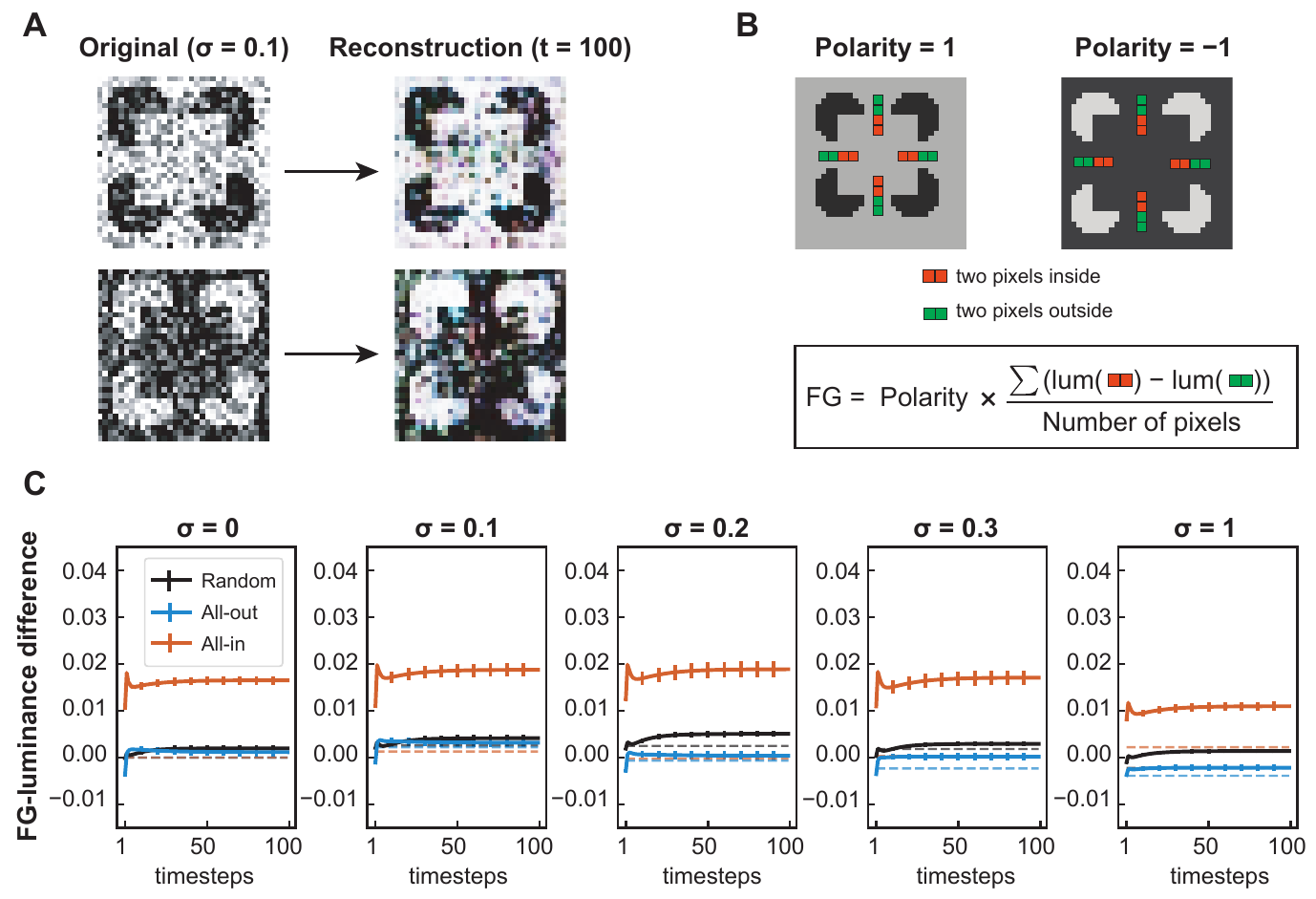}
    \caption{\textbf{Quantification of illusory countour perception in the neural network.} A. Two examples of illusory contour (All-in) stimuli, and their corresponding reconstructions from the network at timestep 100. B. Computation of the ``FG'' value, measuring the figure-ground luminance difference.  C. FG-luminance difference for classes Random, All-out, and All-in over 100 timesteps (zero is absence of illusion, larger values mean more illusion is perceived). Dashed lines denote the FG values computed from original input pictures of the corresponding classes. With clean images, we expect FG on input pictures to be equal to zero as shown in the first plot, while FG values can be bigger or smaller than zero with noisy input images. Results are averaged over 9 networks (distinct random initializations, pretraining and fine-tuning) and error bars represent SEM across networks.}
    \label{fig:fg-value}
\end{figure}

Although the network assigned higher probability of square to the illusory contours than to the control condition, it remains hard to draw the conclusion that the network could really ``see'' illusory contours. It could still be the case that the network is basing its classification decision on other features (e.g. the correctly-oriented corners). A major advantage of the predictive coding model is that we can use its generative feedback pathway to inspect the image reconstructions produced by the model. \textcolor{black}{We inspect the reconstruction of the bottom layer of the network (that is, $d_0(t)=W^b_{1, 0}e_1(t)$).} Figure \ref{fig:fg-value}A displays two examples of reconstructed images at timestep 100. Compared to the original images (on the left) whose noise standard deviation is 0.1, the reconstructed images are denoised by the network, and the illusory contour shapes appear clearer. (Note that this is an emergent property of predictive coding, and not a built-in property of our training scheme: the network's body was not trained with a denoising objective, but only to reconstruct clean natural images). 
To quantify the illusion ``perceived'' by the network, we examined the luminance profile of the reconstructions: for each image we computed a ``Figure-Ground luminance difference'' (FG), as illustrated in Figure \ref{fig:fg-value}B. This statistic measures whether the network perceives an illusory brightness enhancement like humans. Given the expected position of the illusory contour (the position that the square would have occupied if it had been real instead of illusory), along each of the four cardinal axes we took two pixels inside (red in Figure \ref{fig:fg-value}B) and two pixels outside the square (green in Figure \ref{fig:fg-value}B). We computed the average difference between the pixel luminance values inside and outside. A polarity factor (-1 or 1) multiplied this measure, to take into account the different configurations: dark inducers (polarity=1) are expected to produce lighter illusory shapes, light inducers (polarity=-1) to produce darker ones. The constructed FG measure is zero in the original All-in images (since it is measured in the background between the inducers), but should be positive in the image reconstructions whenever an illusory contour is perceived. Figure \ref{fig:fg-value}C shows how this FG value changes over predictive coding iterations, for illusory contours (All-in class) and the other two non-illusory shapes (All-out and Random classes). For the All-in class, after the initial time step the value is consistently higher than zero, and 5 to 10 times higher than the FG value measured for the control (All-out) or the random inducer classes. \textcolor{black}{The corresponding luminance difference is small but reliable (on the order of 0.02, with 1 denoting the full luminance range).} For comparison, the same FG value for the physical square would range from 0.3 to 0.6 (not shown here). \textcolor{black}{Interestingly, as seen above, this perception of illusory contours was still present but somewhat reduced when images were corrupted with extremely high levels of noise ($\sigma=1.0$), a behavior that would be expected in humans. For all other noise-levels that the network was trained to reconstruct}, the results are relatively similar; thus, from here onward we only plot results for a single level of noise (0.1), even though all our conclusions equally apply to all levels.

\subsection{Ablation studies}
\begin{figure}[btp]
    \centering
    \includegraphics{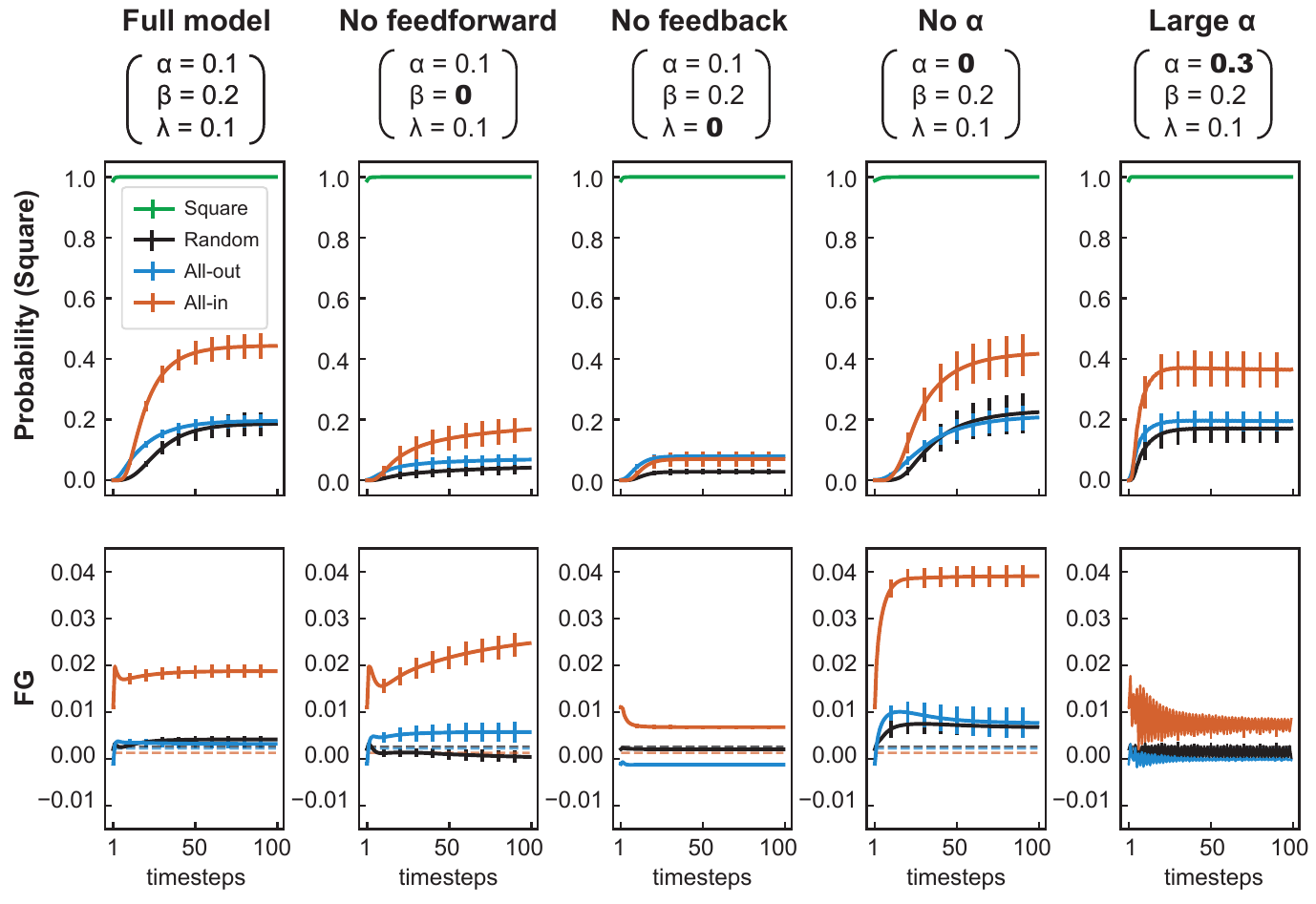}
    \caption{\textbf{Ablation during testing.} The full model results (first column) correspond to the same data already reported in Figures~\ref{fig:decisions} and \ref{fig:fg-value} for $\sigma=0.1$. Probability of square classification is reported in the top row, FG value in the bottom row. Each component of the update equations was set to 0 at test time (with the other parameters at their default value): the feedforward drive term $\beta$ (2nd column), the feedback term $\lambda$ (3rd column), the feedforward error correction term $\alpha$ (4th column). In addition, we tested a larger value of $\alpha$ (last column).}
    \label{fig:ablation_placeholder}
\end{figure}

In order to determine the contribution of the various components of the network to illusory contour perception, we performed systematic ablation studies. For each of the hyperparameters $\alpha$, $\beta$ and $\lambda$ in the update equation~\ref{eq:pc_equation}, we set the parameter to 0 at test time and observed how this affects the network behaviour. Additionally, we also tested a network with a larger value of $\alpha$, keeping all other hyperparameters at their default value. Figure~\ref{fig:ablation_placeholder} shows the results of these ablation experiments. 

As expected, removing feedback ($\lambda$) leads to the complete disappearance of the illusion. This confirms that the generative feedback plays a critical role in producing the illusion. Interestingly, we see that removing the constant feedforward drive $\beta$ (after the initial feedforward activation pass) also seems to diminish the illusory effect, although not completely. It thus appears that both feedforward and feedback contributions are important to see the illusion. Finally, when removing the feedforward error correction term $\alpha$, the square classification probability does not seem to change much; however, the FG values strongly increase, specifically for the All-in condition. This suggests that the network may be seeing the illusion even more strongly. This can be explained by the fact that the role of the feedforward error term $\alpha$ is to update the activations in each layer in order to improve their reconstruction of the layer below. Thus, large FG values (in fact reflecting an imperfect reconstruction of the input image caused by the illusion) are ``corrected'' slightly by the feedforward error correction. When $\alpha$ is set to 0, the network does not correct these large FG values anymore, and is free to ``perceive'' a stronger illusion. To verify that feedforward error correction does indeed suppress the illusory contour perception, we also tested a model with stronger $\alpha$. In this case, both the square classification probability and FG value are visibly decreased for the All-in condition, confirming our hypothesis. 

\textcolor{black}{In Rao and Ballard's original formulation \cite{rao1999predictive}, they treated the two error terms (feedforward and feedback error corrections) together, and did not explore their relative effects. Here we show that only one of these (the feedback error correction) is critical to the perception of illusory contours, but that overall this perception is consistent with the predictive coding framework. However, the network with $\alpha=0$ is essentially equivalent to a regular feedback network, so this suggests that other feedback architectures may also lead to the perception of illusory contours.}

\begin{figure}[htbp]
    \centering
    \includegraphics{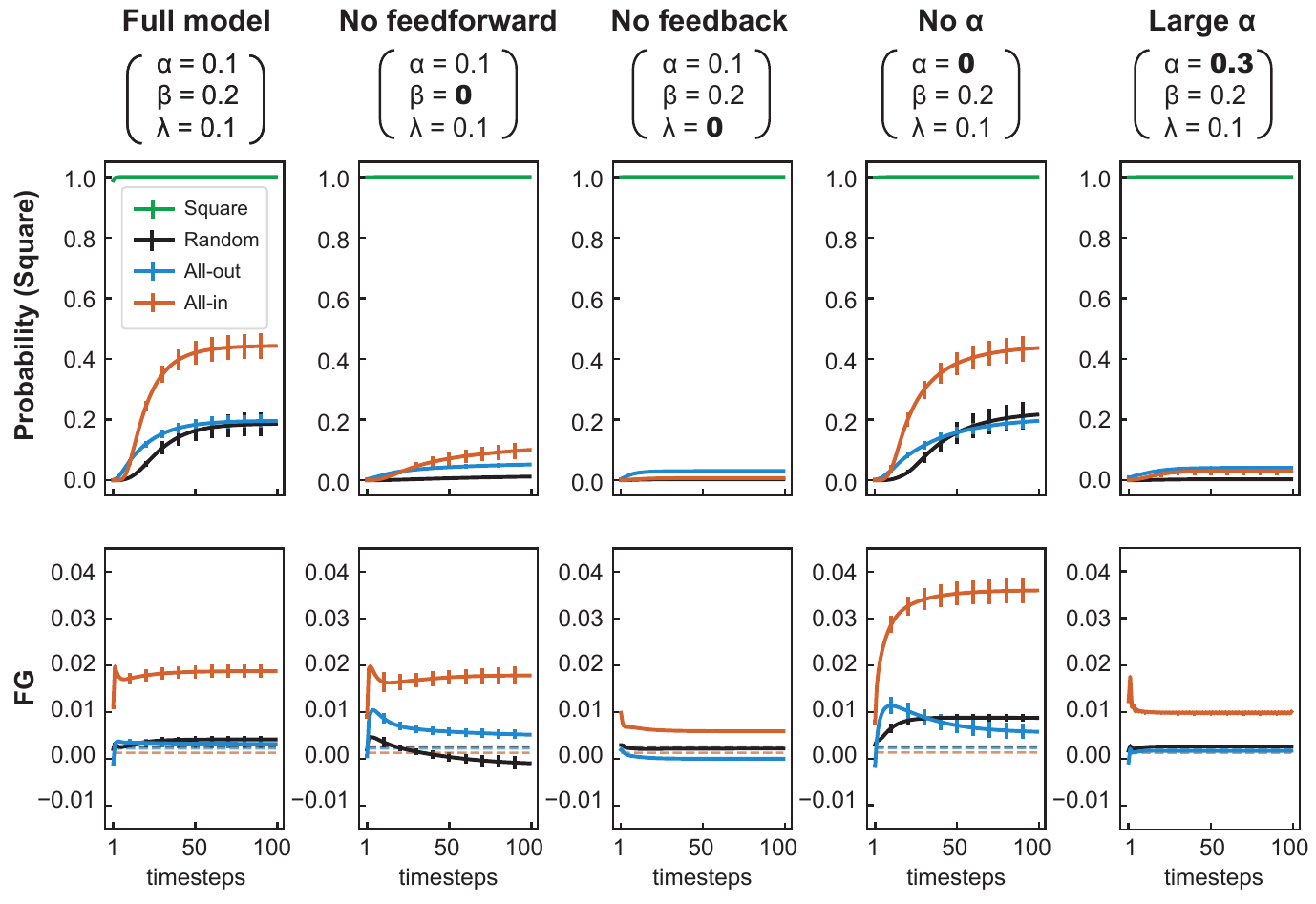}
    \caption{\textbf{Training restricted networks.} These graphs show the behaviour of networks trained from scratch without the various components of the full model. The full model results (first column) correspond to the same data already reported in Figures~\ref{fig:decisions} and \ref{fig:fg-value} for $\sigma=0.1$. Probability of square classification is reported in the top row, FG value in the bottom row. Generally, this produces similar results to just switching off the components at test time. Again, feedback is essential to the perception of the illusion, and feedforward encourages it. On the other hand, $\alpha$ being larger tends to diminish the perception of the illusion.}
    \label{fig:ablation_train}
\end{figure}
\subsection{Training restricted networks} To further explore the effects of the various network components, we also trained networks without each component. The results are shown in Figure \ref{fig:ablation_train}. The difference from the previous ablation studies (Figure \ref{fig:ablation_placeholder}) is that here, rather than being trained with all the components and then having them artificially switched off at test time, here the network ``knows'' during training which components are unavailable, and can potentially learn to compensate for their absence. Thus, where the previous section answered the question ``How does our full trained network behave without each component'', this approach answers the question ``What is the behaviour of the best possible model without each component''. That being said, overall this experiment largely confirms the results of the above ablation tests: that the feedback is essential for the perception of the illusion; that feedforward significantly contributes to this perception; and that larger $\alpha$ (feedforward error correction) diminishes this illusion. One significant difference between Figure \ref{fig:ablation_placeholder} and Figure \ref{fig:ablation_train} is that training with a large $\alpha$ seems to have a stronger deleterious effect than just increasing it at test time. We speculate that this is due to the fact that when the network is trained with large $\alpha$, it implicitly learns to ``trust'' the reconstructions from higher levels more (since they will be corrected faster), and thus learns larger feedback weights.

\begin{figure}[htbp]
    \centering
    \includegraphics[width=\textwidth]{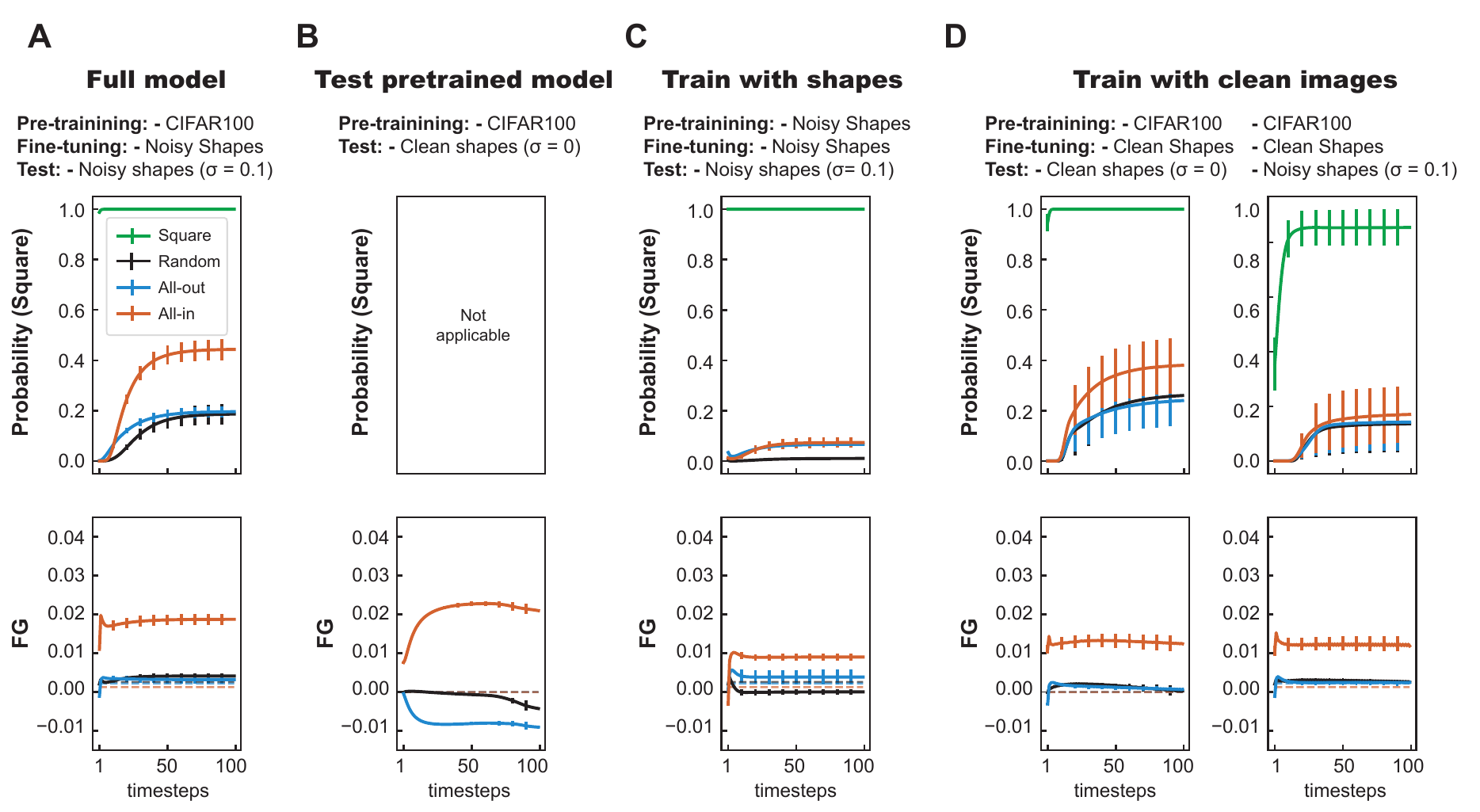}
    \caption{\textbf{Testing the influence of pretraining and finetuning datasets}. A. The full model is the same data already reported in Figures~\ref{fig:decisions}, \ref{fig:fg-value}, \ref{fig:ablation_placeholder} and \ref{fig:ablation_train}, to facilitate comparisons. \textcolor{black}{B. Testing whether networks that have only been pretrained on natural images (no fine-tuning) could perceive illusory contours (this test relies on FG values, as classification probability is not computable for these networks)}  C. Comparing pretraining on CIFAR100 (the Full model) to training directly on the shapes dataset. D. Comparing finetuning on a noisy shape dataset (the Full model) to a clean alternative, i.e. trained and finetuned without noise. The testing is performed without noise (left) or with $\sigma=0.1$ noise (right).}
    \label{fig:dataset}
\end{figure}

\subsection{Pretraining and Finetuning Datasets} \label{natural_scenes}
We also investigated to what extent the specific datasets used for pretraining and/or fine-tuning the network affect its behaviour. Figure \ref{fig:dataset} shows the results of \textcolor{black}{three} comparative tests. \textcolor{black}{In the first test, we took our three pretrained networks (without finetuning on the custom shapes dataset) and tested whether they can ``perceive" illusory contours solely based on the learned statistics of natural images (therefore, this test relies only on the FG figure-ground luminance calculation). In the second test, to further confirm the critical role of \emph{natural images} (instead of other much simpler but task-related images) in illusion perception, we simply removed the pretraining procedure on the CIFAR100 natural image dataset. That is,} we trained our network directly on the shape dataset (for both unsupervised reconstruction pretraining, and supervised classification finetuning). In the third test, we trained without any noise during the finetuning (supervised classification) stage.  

\textcolor{black}{Figures~\ref{fig:dataset}B and C} confirm our hypothesis that pretraining on natural images is necessary and \textcolor{black}{sufficient} for illusion perception. \textcolor{black}{For Figure~\ref{fig:dataset}B, we can only measure the FG values from image reconstructions, but not the classification probabilities, since the pretrained networks were not equipped with a decision head at this stage. Figure C.10-A in the Appendix shows that with the default set of parameters, the pretrained networks can already perceive a certain amount of illusion. Here, to maximize the networks' ability to perceive illusory contours, we applied a different set of parameters with $\beta = 0.1$, $\lambda = 0.2$ and $\alpha = 0$ (since we have observed in our ablation tests above that increasing $\lambda$ and decreasing $\alpha$ could favour the illusion; default parameters are still used elsewhere unless stated otherwise). We also test these networks on clean shapes, because they were only trained on the clean CIFAR100 natural images. From Figure~\ref{fig:dataset}B, it is clear that pretraining on natural images can provide sufficient knowledge about image contour statistics to induce the perceptual illusion. Although the magnitude of the effect is somewhat dependent on the choice of parameters, as detailed above and in our ablation studies (Figures~\ref{fig:ablation_placeholder} and~\ref{fig:ablation_train}), in a separate simulation (Figure S1-B) we show that illusory contours can even arise in a parameter-free alternative training regime (similar to the one described below in ~\ref{vgg_training}). In this case, feed-forward and feedback connections are trained to optimize reconstruction after a single timestep--so the parameters $\alpha$, $\beta$ and $\lambda$ play no role in the outcome. Yet the networks could still perceive illusory contours (Figure S1-B). } \textcolor{black}{Figure~\ref{fig:dataset}C shows that} when the network convolutions are only trained to reconstruct shapes (squares, pacman inducers) instead of natural images, the model processes the All-in and All-out test images similarly. \textcolor{black}{Both results provide} evidence that the perception of illusory contours is an emergent property from the visual system's adaptation to the statistics of the natural world. The \textcolor{black}{third} test investigates the effect of noise (and therefore, uncertainty) during shape classification training. Figure~\ref{fig:dataset}D demonstrates that training without any noise results in a weaker illusory effect for both the square class probability and the FG value. That is, some form of uncertainty or variety in the learned shapes dataset appears helpful; otherwise the network may find a way to perfectly encode or memorise the stimuli, i.e. ``overfit'' the training set.

\subsection{Illusory contour perception in a modern deep network equipped with predictive coding dynamics}  \label{vgg_training}
Finally, we asked whether our approach -- equipping a feedforward network with recurrent predictive coding dynamics -- could be applied to other networks to produce illusory contour perception. Here we used VGG, a much deeper 11-layered network~\cite{simonyan2014very}, which more closely resembles modern state-of-the-art computer vision networks. Previous work from our group demonstrated that the VGG architecture can be equipped with predictive coding dynamics \cite{choksi2020}, and we follow the same approach here.

The basic VGG architecture is a simple feedforward CNN with 8 convolutional layers with 3x3 kernels, max pooling, and 3 fully-connected layers for the head. Although now a few years out of date, it was state-of-the-art in 2014. Compared with the original architecture that was designed for larger images from the Imagenet dataset~\cite{simonyan2014very}, the main difference for our network trained on smaller natural scenes (32x32 pixels) is that the first pooling layer was removed, and the final classification output layer was changed to have 100 instead of 1000 classes. We use a feedforward backbone network where the weights were optimized for natural image classification on the CIFAR100 dataset. The trained model is then augmented with predictive loops consisting of transpose convolution layers, as in the 3-layer model. However for this larger model, the feedback predictions span two feedforward convolutional layers, rather than just one. Thus the inputs to the feedback convolutions are the 2nd, 4th, 6th and 8th convolutional layers (respectively aiming to reconstruct the image input, 2nd, 4th and 6th layers). The weights of these feedback convolutions are optimized for one-step reconstruction over CIFAR100 with a simple mean square error loss (while the feedforward weights remain fixed). Finally the same update equations (Eqs \ref{eq:pc_equation} and \ref{eq:pc_equation2}) are used as with the 3-layer model. This recurrent model is hereafter referred to as PVGG (for ``predictive-VGG''). \textcolor{black}{Notably, the feedback connections in the PVGG model were trained with a different method (already alluded to above), optimizing image reconstruction on the CIFAR100 dataset after a single timestep. Therefore, the parameters $\alpha$, $\beta$ and $\lambda$ play no role in the training. Then, during testing, we selected a set of parameters according to our observations with the smaller model. As we found that the feedforward error correction term $\alpha$ tends to suppress the illusion and the feedback $\lambda$ tends to increase it (Figures \ref{fig:ablation_placeholder} and \ref{fig:ablation_train}), we here set $\alpha=0$, $\beta=0.1, \lambda=0.2$.} 

\begin{figure}[htbp]
    \centering
    \includegraphics{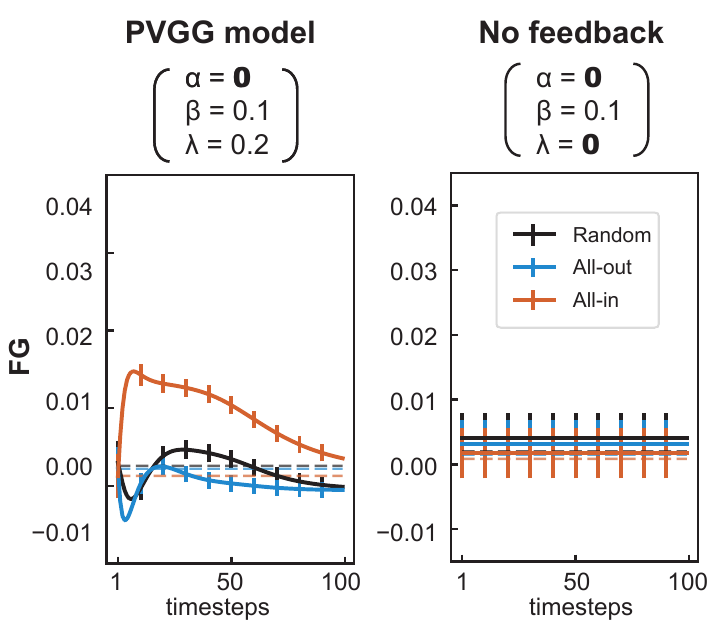}
    \caption{\textbf{FG values in the PVGG network}. Left: The average FG value for the inducer condition is significantly larger than for the two control conditions, demonstrating that the network perceives the illusion. Right: Removing feedback (which here leads to constant activations, since there is no feedforward error correction either) clearly destroys the illusion.}
    \label{fig9:pvgg}
\end{figure}

Since we do not train the model at all on the shapes dataset, we cannot examine the network's classification decisions (since in order to do so, at the very least the classification head would need to be re-trained with the appropriate shape classes). Instead we directly inspected the image reconstructions of the network and calculated the FG values, as illustrated in Figure \ref{fig:fg-value}. Figure \ref{fig9:pvgg} shows the results: it is clear that over timesteps, once again the average FG value for the illusory inducers rapidly becomes much larger than either the random or control conditions. As in previous experiments, we could verify that feedback error correction is critical for this illusory perception: no illusion occurred when we ablated the feedback at test time, i.e. $\lambda=0$ (Figure \ref{fig9:pvgg}, right).

\textcolor{black}{Although this network also appears to perceive the illusion, the long-run dynamics of the network are quite different from the smaller model. While in the smaller model, even after 100 timesteps the illusion is still perceived, in the PVGG model it gradually disappears after some timesteps. In a way, presenting a static stimulus for so many timesteps is somewhat unrealistic - for humans, the input stimulus is constantly changing, either because of environmental fluctuations, or because of saccadic eye movements. Indeed, previous work showed that when illusory contours are perceived by humans in the periphery, they disappear after a few seconds \cite{ramachandran1994perception}. However, the difference between PVGG and the smaller model is interesting, and could be an avenue for further research. A number of key differences between the two models could be at play. Most importantly, the weights of PVGG are not learned `over timesteps' - the feedforward weights are trained just for feedforward classification, and the feedback weights are trained just for a single step reconstruction. Thus, the network never updates its weights after predictive coding updates. On the one hand, this makes it even more remarkable that the updates lead to the perception of the illusion, and is strong evidence that the framework is compatible with illusory contour perception. However, it could also be the reason why the long-term dynamics are somewhat unstable (or non-convergent), unlike in the smaller model. The other significant differences are that PVGG is a much deeper network, and was trained only on natural images, never on the shapes dataset. As such it has much more complex, abstract representations, and it is possible that these representations are not well-calibrated to simple shapes or contours which would not appear in CIFAR images. As a result the network might struggle to maintain stable representations of the illusory contours, and might instead `hallucinate' (or `fill in') what it imagines is missing texture or complexity.}

The above strategy of avoiding re-training or fine-tuning the network has advantages from both an AI and a neuroscience perspective. First, not only does it save the time and power required for training, but it also allows us to experiment with any pretrained feed-forward model -- including models which we would not be able to train ourselves due to computational restrictions. Second, from a neuroscience perspective, the fact that the perception of the illusion can occur in a network which has never seen the shapes dataset before (nor, presumably, any square or pacman shape) demonstrates even more clearly that it results from a combination of feedback connectivity, predictive coding dynamics, and exposure to natural scenes.

\section{Discussion} \label{discussion}

The purpose of this study was to test whether a feedback neural network with brain-inspired recurrent dynamics would perceive illusory contours (Kanizsa squares) in a similar manner to humans. Augmenting a feedforward CNN with predictive coding recurrent dynamics allowed us to (i) analyse explicit classification decisions (square vs. inducers) and, unlike other related work ~\citep{lotter2018neural,Baker2018DeepCN,kim2020neural},  (ii) visualize reconstructed inputs from the model's viewpoint. As reported in a preliminary version of this study recently published in a conference workshop~\citep{pang2000predictive}, we found that, compared to a feedforward baseline, the recurrent dynamics led the network to perceive more illusory contours. Notably, by inspecting the network's reconstructions, we were able to directly visualize the network's internal representation of the stimulus, which provides a much clearer measure of ``illusory perception'' than previous works. We found evidence of modulations of the perceived luminance profiles in the expected direction for illusory shapes, suggesting that the network is truly ``perceiving'' the contours. We extended this analysis and performed systematic ablation studies, both at test time and at training time, and found that the feedback error correction term is essential to the perception of the illusion, while the feedforward error correction term tends to decrease it. Similarly, exploring the datasets used for pretraining and fine-tuning the network revealed that prior exposure to the statistics of natural scenes is a crucial element of illusory contour perception. Finally, we also implemented the predictive coding dynamics in a standard VGG model, and found that the modified PVGG model also exhibited the perception of illusory contours. This suggests that illusory contour perception arises from predictive coding feedback dynamics, independently of the scale of the model. In summary, we provide clear evidence that brain-inspired recurrent dynamics can lead networks to perceive illusory contours like humans.

Although there are intrinsic differences between the human visual system and artificial neural networks (e.g., the global error signals required for learning via back-propagation~\cite{lillicrap2020backpropagation}), we argue that the current findings highlight three key similarities with biological vision. First, both systems may engage similar global processing of illusory contours. In the visual system, Pan et al. reported that illusory contours activate equivalent representations in V4 compared to real contours~\cite{pan2012equivalent}, whereas V1 and V2 differently encode their respective local features. That is, in addition to local processing in early visual regions, there exists a global processing mode whereby illusory inducers form integral contour representations. In a similar way, when presented with illusory contours the current network assigned much higher probability of the ``square'' class than for either Random or All-out control images, though they shared the same local features as the All-in illusory contour images. This indicates that the network also possesses a capability of global processing. Moreover, this global processing primarily results from the feedback connections, since none of the tested feedforward networks could perceive illusory contours (Table \ref{tab:ff_results}). Second, the ``behavioural'' performance (i.e. decision probability) of the network is also consistent with physiological research on illusory contours. Lee et al compared EEG activity for illusory contours and other patterns, and found that the activity for illusory contours is significantly higher than control random stimuli, but still lower than real contours~\cite{lee2001dynamics}. In the current study, the ``Square'' class probabilities assigned by the network after the Softmax layer indicate a similar pattern (see Figure \ref{fig:decisions}). Lastly, at the ``perceptual'' level, we directly checked the internal representation at the first layer of the network (through its generative ``image reconstruction'' pathway). The FG metric suggested that the network perceives a brighter (or darker) illusory shape, consistently with the ``illusory brightness'' reported when humans perceive illusory contours ~\cite{Schumann1901-SCHBZA-3,spillmann1995phenomena,parks2001rock}.

Having designed a biologically inspired architecture, we performed ablation studies to examine how the behaviour of the network depends on its various components. Most importantly, we found that feedback error correction was critical to the network perceiving the illusory contours, which is in agreement with previous results which demonstrated that feedforward networks do not appear to perceive illusory contours~\cite{Baker2018DeepCN}. We also found that removing the feedforward error correction term (one half of the predictive coding update proposed by Rao and Ballard~\cite{rao1999predictive}) seems to enhance the perception of the illusion. This is easily explained as this term serves to correct the representations to minimise reconstruction error of the incoming layer, so the erroneous contour and luminance difference may be ``corrected out'' when this term is present. These ablation experiments allow us to highlight how the two error correction terms, introduced concurrently in Rao and Ballard's paper~\cite{rao1999predictive}, play two distinct roles: the feedforward error correction term encourages the network to accurately represent the stimuli, anchoring it in the ground truth; whereas the feedback error correction tries to explain the input in terms of the network's implicit priors, and in this case causes the network to ``hallucinate'' based on its higher-level representations (for example, closed contours or square shapes). Finally, by scaling the predictive coding updating dynamics to VGG11 and finding the same human-like illusory contours (see Figure \ref{fig9:pvgg}), we provided evidence that these feedback dynamics may induce illusory perception across a large range of deep convolutional network architectures.

In Section \ref{natural_scenes}, we investigated what effect pretraining on natural images has on the network. Previous work suggested that CNNs trained on natural scenes are better models of brain activity~\cite{maheswaranathan2018deep}. Indeed, we saw that training directly on the shapes was not sufficient for the network to assign higher probability of the square class to illusory contours, nor to produce the positive FG values which are the hallmarks of illusory contour perception. Training on natural images encourages the network to learn efficient representations of natural stimuli -- the constraints that arise from exposure to natural scene statistics in the brain have been proposed as a cause of many visual illusions~\cite{eagleman2001visual,gori2016visual}. In other words, pretraining the network on natural images may help the network to learn more human-like representations. We also investigated the effect of training with images corrupted with Gaussian noise versus without. Such data augmentation typically helps in improving generalisation as it forces the network to learn meaningful features from the data instead of rote-memorising individual training examples~\citep{bishop1995training,akbiyik2019data}. We found that networks trained using noisy images perceive stronger and more consistent (as reflected by smaller standard error values) illusory contours.

In summary, by leveraging insights from neuroscience, we designed an original brain-inspired deep learning architecture and thus add to a growing body of literature exploring the cross-pollination of neuroscience and AI. On the one hand, the current study demonstrates that we can effectively use neuroscience principles to design artificial computer vision models, and probe them using classical stimuli and illusions from neuroscience and cognitive science. On the other hand, we also illustrate how modern deep learning techniques can be used as powerful tools in examining our theories of brain function~\cite{kriegeskorte2015deep,cichy2016comparison,vanrullen2017perception,marblestone2016toward}. By building and testing a brain-inspired model, the current study highlights the essential roles of feedback connections~\cite{lee2001dynamics,pak2020top}, predictive coding computation~\cite{notredame2014visual,nour2015perception,raman2016predictive,shipp2016neural}, and prior experience of natural environments~\cite{eagleman2001visual} for the perception of visual illusions. Future work could use the same kind of predictive coding model to test other aspects of the predictive coding theory in neuroscience, such as its tendency to produce oscillatory dynamics~\cite{alamia2019alpha}, or other phenomena observed in human vision, such as ambiguous stimuli and multi-stable perception, or the Gestalt rules of perceptual organization.

\section*{Acknowledgements}

The authors would like to thank Milad Mozafari for providing the implementation for PVGG. This work was funded by an ANITI (Artificial and Natural Intelligence Toulouse Institute) Research Chair to RV (ANR grant ANR-19-PI3A-0004), as well as ANR grants AI-REPS (ANR-18-CE37-0007-01) and OSCI-DEEP (ANR-19-NEUC-0004). ZP is supported by China Scholarship Council (201806620059).

The authors declare no competing financial interests.
\clearpage

\bibliography{mybibfile}
\clearpage

\appendix

\section{Feedforward gradient term normalisation}

During the updates, we scale the feedforward gradient term to normalise the expected size of the gradients. This is done to avoid any dependence on layer or kernel size. Here we derive this scaling factor. During the updates, for layer $n+1$, we update according to the (gradient of the) mean square error of the layer below:
\[\epsilon_{n}=\frac{1}{K}\sum_i(e_n^i-d_n^i)^2\]
where $K$ is the number of elements in layer $n$.
Then looking at an element $j$ of the gradient $\nabla \epsilon_{n+1}$, we can write its partial dervative as the mean of the derivatives of the differences at each location $i$ in layer $n$ below:
\begin{align}
\frac{\partial \epsilon_{n}}{\partial e_{n+1}^j} &= \frac{\partial}{\partial e_{n+1}^j} \frac{1}{K}\sum_i(e_n^i-d_n^i)^2 \\
&= \frac{1}{K}\sum_i\frac{\partial (e_n^i-d_n^i)^2}{\partial e_{n+1}^j}
\end{align}
Now we suppose that the derivatives for each activation are i.i.d. normally distributed around 0 (which would follow from the weights in the deconvolutions being normally distributed):
\begin{equation}\frac{\partial (e_n^i-d_n^i)^2}{\partial e_{n+1}^j}\sim\mathcal{N}(0,\sigma^2)\end{equation}
Then we see that
\begin{equation}
\label{eq4bhavin}
\sum_i^K\frac{\partial (e_n^i-d_n^i)^2}{\partial e_{n+1}^j}\sim\mathcal{N}(0,K\sigma^2)
\end{equation}
so that
\begin{equation}\frac{\partial \epsilon_{n}}{\partial e_{n+1}^j}=\frac{1}{K}\sum_i^K\frac{\partial (e_n^i-d_n^i)^2}{\partial e_{n+1}^j}\sim\mathcal{N}(0,\frac{\sigma^2}{K})\end{equation}
That is to say that the variance is scaled according to the number of activations $K$. In short, when we take an average over samples centred on 0, as the sample grows larger, the average tends to 0. This means that for a large number of activations, we expect the elements in the gradient tensor to be small.

A second effect compounds this problem. In a CNN, most gradients between layers will be zero (since a unit's activation depends only on the activations within its receptive field). So for each $i$ where $d_n^i$ is not in the receptive field of $e_{n+1}^j$, the partial derivative will be 0.  In this case, supposing there are some $C<K$ activations which are dependent (C will be equal to the size of the kernel multiplied by the number of channels), we can replace the factor of K in equation~\ref{eq4bhavin} by $C$ so that
\begin{equation}\sum_i^K\frac{\partial (e_n^i-d_n^ii)^2}{\partial e_{n+1}^j} = \sum_i^C\frac{\partial (e_n^i-d_n^i)^2}{\partial e_{n+1}^j}\sim\mathcal{N}(0,C\sigma^2)\end{equation}
Then we in fact have that
\begin{equation}\frac{\partial \epsilon_{n}}{\partial e_{n+1}^j}=\frac{1}{K}\sum_i^C\frac{\partial (e_n^i-d_n^i)^2}{\partial e_{n+1}^j}\sim\mathcal{N}(0,\frac{C\sigma^2}{K^2})\end{equation}
for this $C<K$. That is, the gradients shrink even more rapidly as K grows. Thus, to counteract these two effects, we normalise the gradient term by $K/\sqrt{C}$ (where $K$ and $C$ can be different for each layer). Specifically, $K$ is equal to the total number of activations in layer $n$, and $C$ is equal to the product of the kernel size and number of channels in layer $n$.

\section{Relationship with Rao and Ballard's formulation}

This work builds directly from the Rao and Ballard's formulation of predictive coding. However, in our case we use automatic differentiation to minimize the prediction error by taking iterative steps in the opposite direction to the error gradient $\nabla\epsilon$. On the other hand, Rao and Ballard's method propagates the error residual $\epsilon$ through the transposed feedback weight matrix $U$. Here we demonstrate the equivalence of the two error minimization strategies. We first note that, by rearranging Equation 1 from the main text, the feedback error term becomes $\lambda(d_n(t) - e_n(t))$ which is exactly the same as the term $r^{td}-r$ (in their notation) in Equation 7 of Rao and Ballard (Page 86).

We next show that the feedforward gradient term is equal to the pixel-wise error passed through the matrix of top-down connection weights. We define the error in layer $n$ as:
\[\epsilon_{n}=\frac{1}{K}\sum_i(e_n^i-d_n^i)^2\]
Then we take the gradient of the error with respect to the activations in layer $n+1$ above:
\begin{align}
	\nabla\epsilon_{n} &= \begin{bmatrix}
							\frac{\partial\epsilon}{\partial e_{n+1}^1} \\
							\vdots \\
							\frac{\partial\epsilon}{\partial e_{n+1}^K}
					\end{bmatrix}
\end{align}
We use this to update the activations of the upper layer. Then the element of the gradient at position $j$ is
\begin{align}
	(\nabla\epsilon_{n})_j & = \frac{\partial\epsilon_{n}}{\partial e_{n+1}^j} \\
					 & = \frac{\partial}{\partial e_{n+1}^j} \frac{1}{K}\sum_i(e_n^i-d_n^i)^2 \\
					 & = \frac{1}{K}\sum_i\frac{\partial (e_n^i-d_n^i)^2}{\partial e_{n+1}^j} \\
					 & = \frac{1}{K}\sum_i 2(e_n^i-d_n^i) \frac{\partial e_n^i}{\partial e_{n+1}^j}
\end{align}
Then $\frac{\partial e_n^i}{\partial e_{n+1}^j}$ is exactly the weight of the top-down connection between $e_n^i$ and $e_{n+1}^j$. That is, if $U$ denotes the matrix transformation from the upper layer to the lower layer, $\frac{\partial e_n^i}{\partial e_{n+1}^j}=U_{ji}$, then
\begin{align}
	(\nabla\epsilon_{n})_j = \frac{1}{K}\sum_i 2(e_n^i-d_n^i) U_{ji}
\end{align}
Thus we see the close relationship to the Rao and Ballard method, as this is exactly passing the pixel-wise (or activation-wise) error through the matrix of top-down connection weights. One subtle difference to note is that Rao and Ballard assume symmetric bottom-up and top-down connection weights, such that the error can be propagated directly through the feedforward connections. Here we do not in fact use tied weights (or the transpose weight matrix) - the forward and backward weights are learned simultaneously; but since the gradient is computed directly, the error minimization can be obtained without assuming symmetric connection weights. The only other differences are the constant K (a result of us passing the average error, not the sum, although we correct this with our normalisation), and the factor of 2 (which is somewhat arbitrary given that we use a hyperparameter to scale the gradient anyway). A fuller derivation of the equivalence of the two formulations can be found in Choksi, Mozafari et al. \cite{choksi2021predify}.

\section{Influence of number of \emph{timesteps}}
As we stated in the main paper, we chose 10 timesteps  to produce meaningful dynamic trajectories while remaining within our computational constraints. However, we show here that we can still see reasonable illusory perception results when training for other numbers of timesteps. Figure \ref{fig:decisions}A shows that training with either fewer or more timesteps also leads to illusion perception. The various hyperparameters (including number of timesteps) are highly interdependent, so changing one is likely to lead to the others being suboptimal. Due to computational limitations, we were not able to, during the review process, search for optimal hyperparameters for other choices for number of timesteps; we just slightly tweaked some hyperparameters based on our understanding of the network to favor the emergence of the illusion. Panel A reveals that the illusion can still be seen (even if it is slightly weaker) when training for different numbers of timesteps. Forthcoming work will investigate the effect of the hyperparameters, and their dependencies on the training timesteps, more fully.

To further test this claim, we also used a parameter-free alternative training regime, exactly as with the PVGG model. In this approach, the network's feedforward weights were optimized for natural image classification on the CIFAR100 dataset. After augmenting the network with predictive loops, the feedback weights are optimised for one-step reconstruction over CIFAR100 with a simple mean square error loss (while the feedforward weights remain fixed). In this case, feed-forward and feedback connections are trained to optimize reconstruction after a single timestep--so the parameters $\alpha$, $\beta$ and $\lambda$ play no role in the training, and at test time the network dynamics can subsequently be left to unfold over multiple time steps. The trained networks were tested on the shapes dataset with proper parameter configuration. The results (Figure~\ref{fig:decisions}B) show that the networks could still perceive the illusion.

\begin{figure}[htbp]
    \centering
    \includegraphics{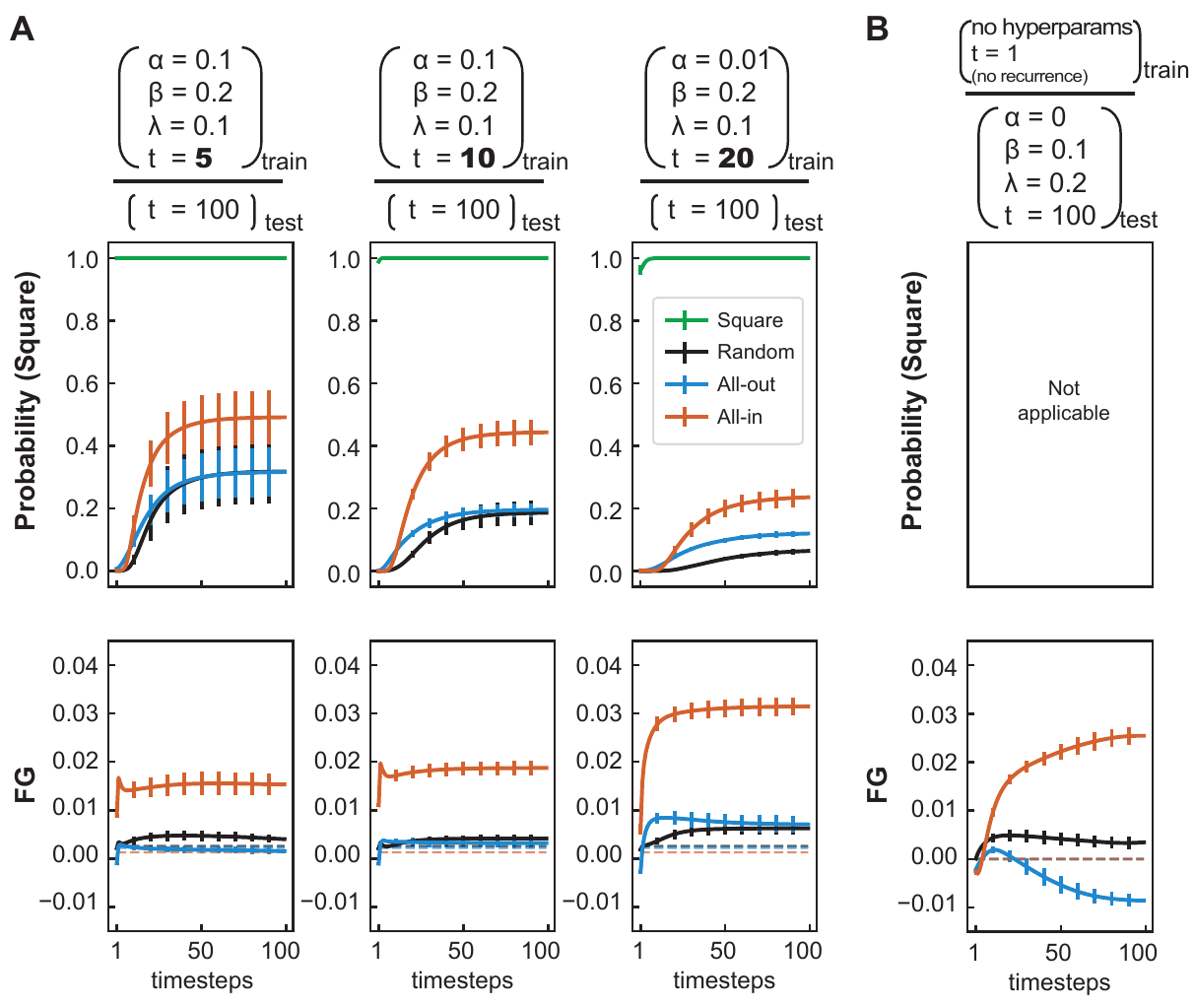}
    \caption{\textbf{Influence of training timesteps}. A. Training networks with different timesteps. To produce a reasonable amount of illusion, the $\alpha$ value was adjusted when timesteps = 20. During testing, the same set of parameters were used respectively except that the number of timesteps was set to 100. B. A new training regime was used where no parameters were applied. Since networks were only trained on CIFAR100 dataset and were not finetuned on shapes dataset, only FG results are presented to evaluate illusion perception.} 
    \label{fig:decisions}
\end{figure}

\end{document}